\documentclass{article}

\usepackage[preprint]{corl_2026} 

\usepackage{enumitem}
\usepackage{soul}
\usepackage{amsfonts}
\usepackage{amsmath}
\usepackage{booktabs}
\usepackage{graphicx}
\usepackage{caption}
\usepackage{algorithm}
\usepackage{algpseudocode}
\usepackage{wrapfig}
\usepackage{makecell}
\usepackage[table]{xcolor}
\usepackage{twemojis}
\usepackage{array}

\usepackage[toc,page,header]{appendix}
\usepackage{minitoc}

\newcommand{\algphase}[1]{%
    \vspace{0.3em}%
    \Statex \hspace{\algorithmicindent}%
    \raisebox{-0.15ex}{\textbf{\textsc{#1}}}%
    \vspace{0.4em}%
}

\newlist{compactenumerate}{enumerate}{1}
\setlist[compactenumerate]{nosep, topsep=0pt, left=3pt, label=\arabic*.}

\newlist{compactitemize}{itemize}{1}
\setlist[compactitemize]{nosep, topsep=0pt, left=3pt, label=\textbullet,}

\title{\raisebox{-0.25em}{\resizebox{!}{1.4em}{\twemoji{telescope}}}\;%
       \underline{Scout}ing the Dynamics Gap: Test-Time Policy Adaptation via Action-Outcome Feedback}

\author{
  Yishu Li*$^{1}$, Liyuan Geng*$^{1}$, Xinyi Mao*$^{2}$, Amber Li$^{1}$, David Held$^{1}$ \\
  $^{1}$Robotics Institute, Carnegie Mellon University \\ $^{2}$Computer Science and Technology, Tsinghua University\\
}

\begin{document}
\doparttoc 
\faketableofcontents
\maketitle


\begin{abstract}
    While pretrained robotic policies exhibit impressive capabilities in controlled environments, 
    unobserved physical properties and dynamics  
    require these policies to rapidly adapt during deployment. 
    Existing test-time adaptation methods typically rely on sparse scalar rewards, failing to exploit the rich geometric and dynamic feedback from the environment during physical interaction. 
    To address this challenge, we propose SCOUT, a dynamics-aware meta-learning framework that enables manipulation policies to rapidly adapt by continuously revising their internal beliefs about environment dynamics. Our approach couples an action-prediction policy with a forward dynamics model via a shared belief latent space. During meta-training, an inner loop updates this shared belief latent by minimizing the dynamics prediction error against the observed action outcome, while the outer loop optimizes the network for action selection. At deployment, this structure allows the agent to infer and adapt to unknown physical dynamics on the fly. By updating its latent belief based on action-outcome mismatches, the policy automatically adapts without risking catastrophic forgetting. We demonstrate that SCOUT significantly accelerates online adaptation across simulated manipulation benchmarks and achieves robust sim-to-real transfer in the real world.  Our project website is available at: \href{https://liy1shu.github.io/SCOUT/}{https://liy1shu.github.io/SCOUT/}.
\end{abstract}


\section{Introduction}
    
While pretrained robotic policies have demonstrated impressive capabilities in controlled settings, they are rarely perfect when deployed in the real world. Variations in object physical properties, environmental dynamics, and out-of-distribution states require these policies to adapt on the fly. In humans, navigating this physical ambiguity is remarkably seamless. Cognitive psychology and neuroscience suggest that humans achieve this via an internal ``intuitive physics engine'' that runs rapid, approximate forward simulations of the environment \cite{battaglia2013simulation}. As we interact with the world, our brains extract abstract, invariant representations of hidden physical properties, such as mass or friction, that readily generalize across vastly different scenarios \cite{schwettmann2019invariant}. Crucially, when an object behaves unexpectedly, the mismatch between our generative mental predictions and sensory reality serves as a powerful signal to immediately update our internal beliefs \cite{gershman2019generative}. 

\begin{center}
    \includegraphics[width=\textwidth]{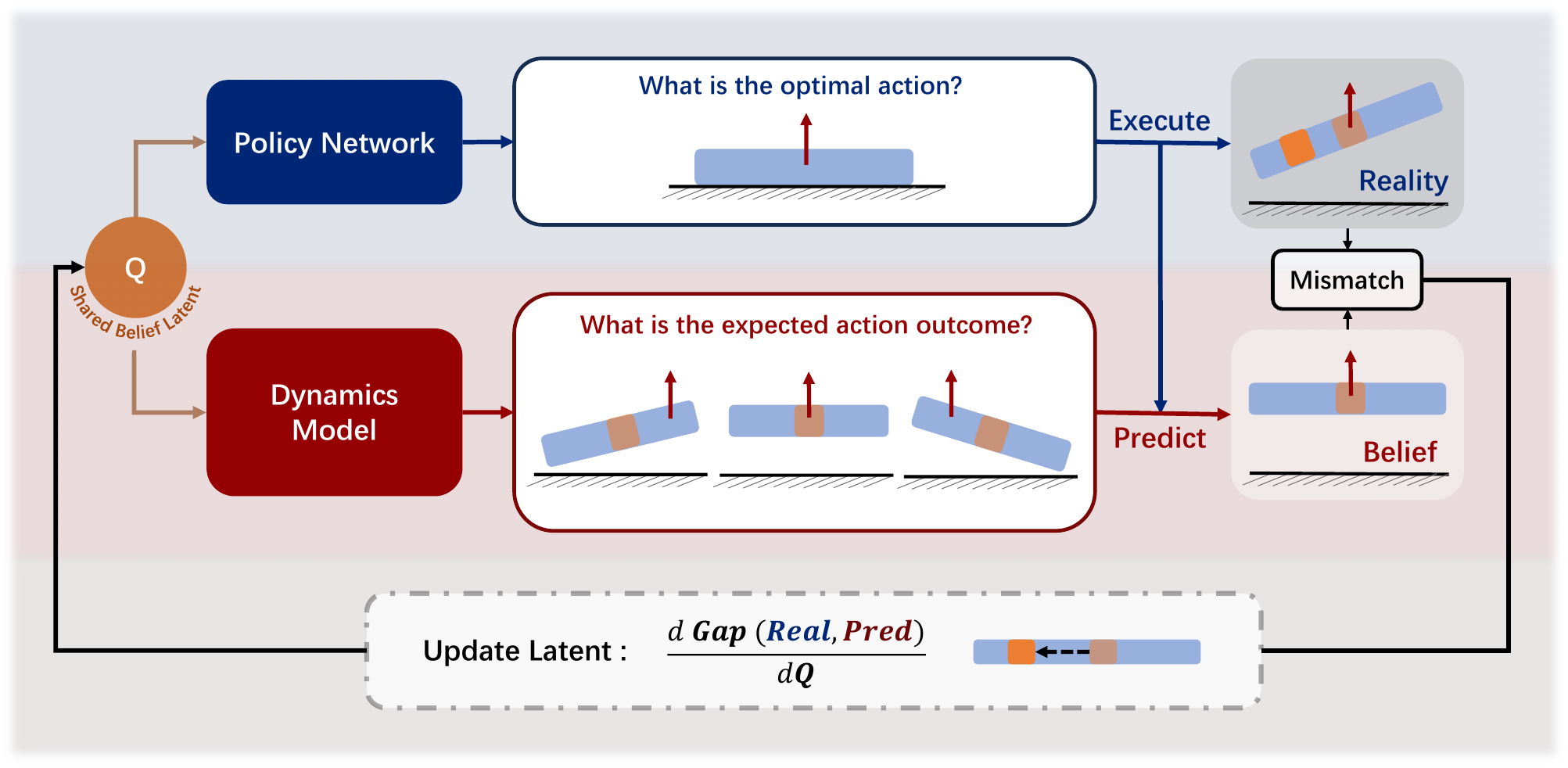} 
    \captionof{figure}{We present \textbf{SCOUT}, which uses a \textbf{\underline{S}}hared belief latent to \textbf{\underline{C}}ouple the policy with dynamics predictions to enable \textbf{\underline{O}}utcome-driven policy \textbf{\underline{U}}pdates at \textbf{\underline{T}}est-time, enabling manipulation policies to adapt to different environment dynamics on the fly.
} 
    \vspace{-2pt}
    \label{fig:teaser}
\end{center}

In contrast, existing test-time training (TTT) and reinforcement learning (RL) methods in robotics struggle to utilize this rich post-action feedback. They are typically limited by a reliance on scalar feedback, such as sparse or hand-engineered dense rewards. Designing continuous reward functions is notoriously difficult and labor-intensive. Furthermore, even when such reward engineering is achieved, a simple scalar signal inevitably discards crucial geometric and dynamic information. For example, if a robot attempts to lift an uneven bar, the specific way the bar tilts provides immediate information about whether the center of mass lies to the left or right, and by exactly how much. 
Relying on a simple scalar reward reduces this crucial geometric data into a generic negative penalty, leaving the system blind as to whether it should adjust its grip to the left or the right.
While we would ideally use the visual and physical mismatch from such interactions to update the policy, mapping raw observational feedback to action corrections remains a significantly hard problem. Because existing robotic methods lack a principled mechanism to extract invariant physical beliefs and translate prediction errors into actionable updates, they struggle to adapt to objects and environments with hidden dynamics and physical properties during deployment.

To address this challenge, 
we propose SCOUT (Fig.~\ref{fig:teaser}), a dynamics-aware meta-learning framework that enables a policy to learn directly from its own failures. Our key insight is that a robust policy inherently relies on an internal ``belief'' about the environment dynamics, such as the location of a door link or the distribution of mass within an object. We propose an architecture in which this belief is shared in a latent space between two modules: a policy that predicts the  manipulation action, and a dynamics model that predicts the outcome of an action. When the robot interacts with the environment, the difference between the predicted outcome and the actual observation provides a rich supervision signal. Using a meta-learning strategy, our inner loop updates the shared latent space using the dynamics prediction loss, while the outer loop supervises the network using an imitation learning loss.  
With this strategy, the model learns weights and an initial belief optimized for test-time adaptation; as the dynamics model revises the shared belief based on outcome prediction errors, the policy automatically adapts accordingly.


We demonstrate the effectiveness of our approach across multiple environments that fundamentally require on-the-fly disambiguation. Through both simulation and real-world experiments, we show that our system can seamlessly adapt to environments and objects with unobserved physics. The contributions of our paper include:

\begin{compactenumerate}[label={(\arabic*)}]
    \item A novel dynamics-aware architecture that links a policy and a dynamics model via a shared belief latent.
    \item A meta-learning training strategy that enables rapid test-time adaptation driven by dynamics prediction errors.
    \item Simulated and real-world experiments demonstrating robust adaptation to environments and objects with hidden physical properties and dynamics.
\end{compactenumerate}
\section{Problem Statement and Assumptions}

We formulate the problem as policy learning under a partially observable Markov decision process (POMDP) subject to unknown or unobserved dynamics. In this setting, the environment's behavior depends on latent physical properties that are not directly observable but are critical for accomplishing the task. 
Formally, we consider a family of POMDPs indexed by a latent variable $z \in \mathcal{Z}$ that specifies the environment's unobserved dynamics properties. Each instance is defined as
$\mathcal{M}_{z} = (\mathcal{S}, \mathcal{A}, \mathcal{O}, \mathcal{T}_{z}, \Omega)$, where the agent observes $o_t \in \mathcal{O}$, takes action $a_t \in \mathcal{A}$, and the state evolves as $s_{t+1} \sim \mathcal{T}_{z}(s_t, a_t), o_t \sim \Omega(s_t)$ with an observation function $\Omega$.  Because the latent $z$ is not directly observed, the agent must (implicitly or explicitly) identify it from observations gathered during interaction. 

\section{Related Works}

\paragraph{Test-time training and meta-learning.}
Prior methods for visual test-time training (TTT) adapt to visual distribution shifts via unsupervised appearance objectives like entropy minimization~\citep{sun2020ttt, wang2020tent, niu2022efficient, gong2022note, gandelsman2022test}. Conversely, gradient-based meta-learning~\citep{finn2017maml, nichol2018reptile} and meta-RL~\citep{duan2016rl2, wang2016learning, rakelly2019pearl, zintgraf2020varibad, fakoor2020meta} explicitly optimize for rapid online adaptation. However, these approaches typically drive adaptation using sparse rewards or coarse trajectory statistics, which is notably less efficient and informative than our use of action-outcome dynamics prediction errors. The closest methods differ from ours in \emph{which} test-time signal drives adaptation, and in \emph{what and how} that signal updates. PEARL~\citep{rakelly2019pearl} and VariBAD~\citep{zintgraf2020varibad} infer a task latent with an amortized learned encoder in a forward pass rather than through gradient updates without explicit error signal to correct the latent when it is wrong. CAVIA~\citep{zintgraf2019cavia} adapts a low-dimensional context vector by gradient descent as we do, but descends the task objective itself, which requires reward or action labels at test time rather than self-supervised action-outcome error. CaDM~\citep{lee2020context} learns a context-conditioned dynamics model for model-based RL, orthogonal to adapting a reactive policy. Closer to our signal, PAD~\citep{hansen2020pad} also adapts during deployment from a self-supervised inverse-dynamics loss, but it has no explicit representation of the hidden environment property: the loss updates the policy's feature extractor, so a change in physics is absorbed as a change in perception rather than estimated as a belief the policy can act on. Concurrent work WorldAgen~\citep{wan2026worldagen} also uses forward prediction error, but finetunes LoRA weights of a backbone shared with the policy after a separate exploration stage, so the policy benefits indirectly without an explicit mechanism that aligns the update with what it needs. Distinct from all of the above, SCOUT updates a belief latent shared by the policy and the dynamics model from self-supervised action-outcome error, meta-learned so that this error translates directly into better actions.  

\paragraph{In-context learning, interactive perception, and history-aware policies.} 
To handle unobserved physics without test-time gradient updates, agents often infer hidden parameters directly from their interaction history. Rooted in interactive perception~\citep{bohg2017interactive, katz2014interactive, weng2024interactive}, modern approaches frame history-dependent adaptation as online system identification~\citep{yu2017uposi, peng2018sim, andrychowicz2020learning, lee2020learning, kumar2021rma}, sequence prompting~\citep{zhang2025dynamics}, or history-aware action generation~\citep{zhao2023learning, li2024flowbothd} and verification~\citep{lihave}. While powerful, these methods typically rely on implicit sequence aggregation without an explicit supervision signal to correct their internal beliefs when the predicted actions fail to achieve the desired state.

\paragraph{Predictive models and physical dynamics.}
Self-supervised world models learn forward environment dynamics for representation learning~\citep{ha2018world, hafner2019planet, hafner2019dreamer} and model predictive control~\citep{finn2017deep, ebert2018visual}. While pretraining such models improves real-world transfer~\citep{wu2023daydreamer, levy2026simulation}, they are typically used for forward planning or offline distillation~\citep{hansen2022tdmpc} rather than for online adaptation. Consequently, these models are rarely used to continuously update the internal representations of deployed reactive policies at test time, and when they are, the coupling to the policy is left implicit..

Our work bridges the gaps across these three lines. We utilize a meta-learning framework for fast execution-time adaptation.  Rather than relying on scalar rewards, we drive the inner-loop TTT update using dynamics prediction errors. By explicitly coupling the policy with a forward dynamics model via a shared belief latent, we convert action-outcome feedback into test-time policy updates.

\section{Method}

\begin{wrapfigure}{r}{0.6\textwidth}
    \centering
    \vspace{-10pt}
    \includegraphics[width=\linewidth]{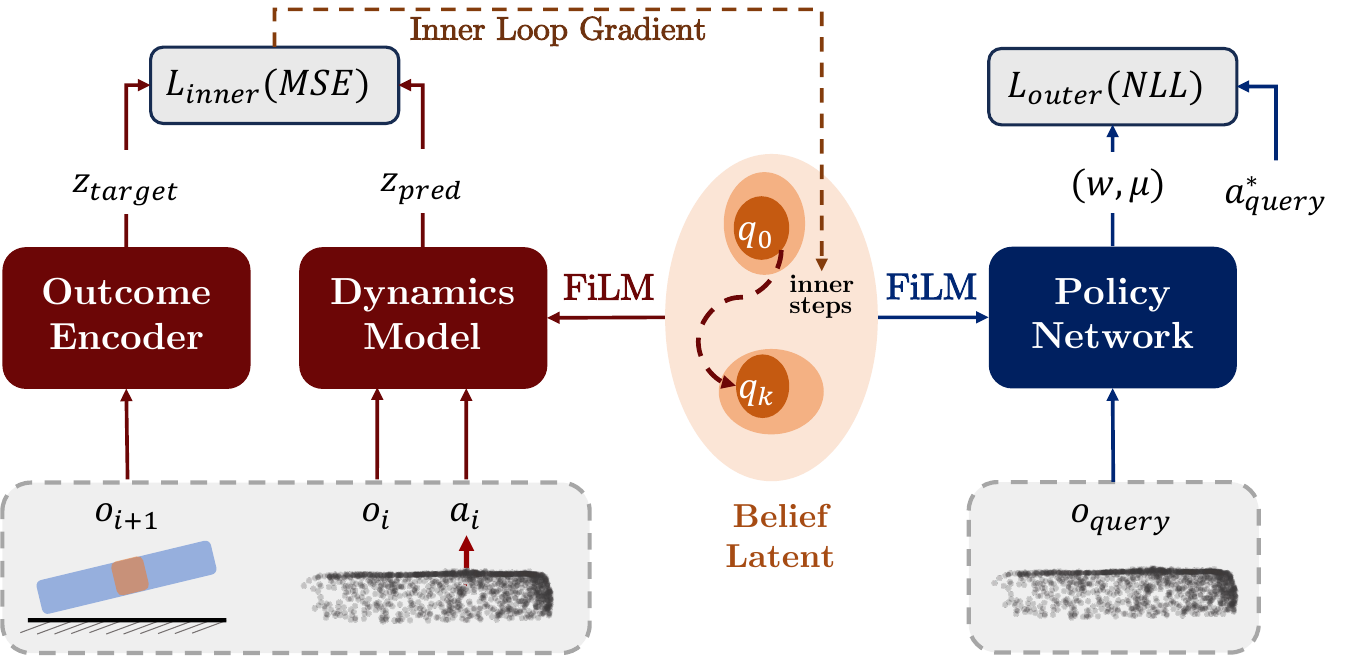} 
    \caption{\textbf{Overview of our test-time adaptation framework.} In the inner loop, the initial belief $q_0$ is adapted to $q_k$ using past interactions by minimizing the latent dynamics prediction error ($L_{\text{inner}}$) while keeping all network weights frozen. In the outer loop, the adapted belief $q_k$ modulates the policy head via FiLM to predict the action distribution. The entire framework, including $q_0$, is meta-trained end-to-end by backpropagating the policy loss ($L_{\text{outer}}$) through the inner loop updates.
    }
    \label{fig:method}
    \vspace{-10pt}
\end{wrapfigure}

Based on the above motivation, we propose a test-time adaptation framework that couples a policy with a predictive dynamics model via a shared belief latent. By optimizing this shared representation through an inner loop of dynamics prediction within a meta-learning framework, we provide a principled mechanism to translate  dynamics prediction errors into actionable policy updates.
An overview of our framework is shown in Fig.~\ref{fig:method}.

\subsection{Shared Belief Latent for Policy and Dynamics}

We denote the agent’s current observation at time step $t$ as $o_t$.
The agent processes this observation to predict a continuous action representation $a \in \mathbb{R}^{d_a}$, which serves as the parameterization for a task-specific target pose. 
To capture the underlying, ambiguous physical properties of the object, our model infers a shared belief latent vector $q \in \mathbb{R}^{d}$. This latent variable intuitively contains the essential information needed to infer the object's dynamics and serves as the critical bridge between the dynamics model and the action-prediction policy.

\textbf{Forward Dynamics Model:} To ground the belief $q$ in physical reality, we introduce a forward dynamics model. 
Following the execution of an action $a_t$ given the initial observation $o_t$, the agent receives the resulting post-action observation (action outcome), denoted as $o_{t+1}$. Depending on the specific task, $o_{t+1}$ can be the raw subsequent point cloud or a per-point flow field (refer to Appendix~\ref{sec:representation} for details).


The Outcome Encoder $E$ maps the action outcome $o_{t+1}$ into a compact target latent space $Z$: $z_{\text{target}, t} = E(o_{t+1})$. Projecting the outcome into a lower-dimensional latent space makes the subsequent inner-loop optimization significantly more stable and computationally efficient compared to directly predicting high-dimensional raw observations.


The Forward Dynamics Model $f$ is a per-point transformer that attempts to predict this exact outcome latent $z_{target}$ prior to action execution. Conditioned on $q$ through FiLM conditioning, it processes $o_t$ and $a_t$ to predict the expected outcome in the shared latent space $Z$: $z_{\text{pred}, t} = f(q, a_t, o_t)$. 
Notably, $f$ predicts the action outcome latent instead of the raw high-dimensional outcome $o_{t+1}$.

\textbf{Policy Network:} Conditioned on the shared belief $q$ and a query state observation $o_{query}$, the policy head employs a Gaussian Mixture Model (GMM) to predict a multimodal distribution over the continuous action space. The belief latent $q$ is injected into the policy through Feature-wise Linear Modulation (FiLM) conditioned transformer blocks. 
This allows the dynamically adapting belief to effectively modulate the extracted features and guide the predicted action distribution.
It outputs a set of mixture weights $w$ and Gaussian means $\mu$:
$(w, \mu) = \text{GMM}(q, o_{query})$. 
More details about the network architectures can be found in Appendices~\ref{sec:policy_network} and  \ref{sec:outcome_encoder}.

\subsection{Test-Time Adaptation via Dynamics Feedback (Inner Loop)}

During deployment, the agent accumulates a history of interactions $H = \{(o_t, a_t, o_{t+1})\}_{t=1}^T$. We utilize Test-Time Training (TTT) to dynamically update the shared belief from a learnable initialization $q_0$ to fit the latent dynamics of the current environment.

In the ``inner loop'' of our method, $q$ is adapted over the history steps by minimizing the Mean Squared Error (MSE) between the Dynamics Model's latent predictions $z_{pred, t}$ and the actual action-outcome latent targets encoded by the Outcome Encoder $z_{target, t}$. Crucially, during test-time deployment, all network weights (including the Policy Network, Outcome Encoder, and Dynamics Model) remain frozen. The only parameter updated is the shared belief latent $q$. Therefore, the loss is explicitly formulated as a function of $q$:
$L_{\text{inner}}(q) = \frac{1}{T} \sum_{t=1}^T \left\| f(q, a_t, o_t) - E(o_{t+1}) \right\|_2^2$.
Starting from the meta-learned initialization $q_0$, we perform $k$ steps of gradient descent on this objective to yield the adapted latent $q_k$. The update rule at step $j$ is given by:
$q_{j+1} = q_j - \alpha \nabla_{q_j} L_{\text{inner}}(q_j)$,
where $\alpha$ is the inner-loop learning rate. Because $q_k$ is optimized to accurately predict outcomes across past interactions, it effectively encapsulates a refined belief regarding the underlying physical properties without risking catastrophic forgetting of the pre-trained policy weights. This adapted $q_k$ is then passed as the conditioning variable to the downstream Policy Network. We provide a structural summary of this process in Figure \ref{fig:method}.

\subsection{Meta-Learning the Shared Representation (Outer Loop)}

\begin{wrapfigure}[26]{R}{0.55\textwidth}
  \vspace{-5pt} 
  \begin{minipage}{\linewidth}
    \makeatletter\def\@captype{algorithm}\makeatother
    
    \hrule height 0.8pt
    \vspace{3pt}
    
    \caption{Dynamics-Aware Policy Adaptation}
    \label{alg:meta_tta}
    
    \vspace{-5pt}
    \hrule height 0.4pt
    \vspace{3pt}
    
    \begin{algorithmic}[1]
    \Require Data $\mathcal{D}_{\mathrm{offline}}$, step sizes $\alpha, \beta$, inner steps $K$
    \Ensure Meta-learned $\theta$ and initial belief $q_0$
    
    \State Initialize $\theta = \{\theta_{\pi}, \theta_f, \theta_E\}$ and $q_0$
    
    \While{not converged}
        \State Sample trajectory batch $\mathcal{B} \sim \mathcal{D}_{\mathrm{offline}}$
        \State $\mathcal{L}_{\mathrm{meta}} \gets 0$
    
        \ForAll{$\tau \in \mathcal{B}$}
            \State Sample history $H = \{(o_t, a_t, o_{t+1})\}_{t=1}^{t}$
            \State Sample query pair $(o_{\mathrm{query}}, a_{\mathrm{query}}^*)$
    
            \algphase{Inner adaptation}
    
            \For{$j = 0, \ldots, K-1$}
                \State $z_t \gets E_{\theta_E}(o_{t+1})$; $\hat{z}_t \gets f_{\theta_f}(q_{j}, a_t, o_t)$
                \State $\mathcal{L}_{\mathrm{inner}} \gets \frac{1}{|H|} \sum_{t \in H} \left\| \hat{z}_t - z_t \right\|_2^2$
                \State $q_{j+1} \gets q_{j} - \alpha \nabla_{q_{j}} \mathcal{L}_{\mathrm{inner}}$
            \EndFor
    
            \algphase{Outer meta-objective}
            \State $(w,\mu) \gets \pi_{\theta_{\pi}}(q_{K}, o_{\mathrm{query}})$
            \State $\mathcal{L}_{\mathrm{meta}} \mathrel{-=} \log p_{\mathrm{GMM}}(a_{\mathrm{query}}^* \mid w,\mu)$
        \EndFor
    
        \algphase{Meta-update}
        \State $\theta \gets \theta - \beta \nabla_{\theta} \mathcal{L}_{\mathrm{meta}}$; $q_0 \gets q_0 - \beta \nabla_{q_0} \mathcal{L}_{\mathrm{meta}}$
    \EndWhile
    \end{algorithmic}
    
    \vspace{3pt}
    \hrule height 0.8pt
    
  \end{minipage}
  \vspace{-10pt}
\end{wrapfigure}

We train the entire pipeline end-to-end using a meta-learning framework. The overall flow of the training process is shown in Algorithm \ref{alg:meta_tta}. While the inner loop adapts $q$ using historical interactions, the outer loop evaluates the adapted $q_k$ on a future query step ($o_{\text{query}}, a_{\text{query}}$) from the same trajectory. 
The outer loop objective is a Negative Log-Likelihood (NLL) loss, which maximizes the probability of the ground-truth query action $a_{\text{query}}$ under the predicted GMM distribution. Formally, for a mixture of $K$ components with predicted weights $w_k$ and means $\mu_k$, the outer loss is defined as:
$
L_{\text{outer}} = - \log \left( \sum_{k=1}^K w_k \cdot \mathcal{N}(a_{\text{query}} \mid \mu_k, \Sigma) \right)
$
where $\Sigma$ represents the covariance matrix of the Gaussian distribution.

To enable this meta-learning process, the outer loop evaluates the policy's performance using the adapted belief $q_k$. By differentiating this outer objective $L_{\text{outer}}$ with respect to the network weights and the initial latent $q_0$ through the sequence of inner-loop updates, we explicitly optimize the model's ability to adapt during test time. 
The fundamental goal of this coupled training scheme is to learn a set of model weights and an initial belief $q_0$ that are optimized for fast adaptation. Rather than merely learning static features, this ensures the entire pipeline learns to rapidly update its internal dynamics belief from post-action feedback at test time.

Note that the latent dynamics loss $L_{\text{inner}}(q)$ can be prone to mode collapse, in which the dynamics model and the outcome encoder always output the same constant embedding, which would minimize the inner loop loss and yield zero gradients, preventing any meaningful belief updates during adaptation.  For tasks with high ambiguity, the outer loop loss prevents this from happening, since the outer-loop loss will drive the dynamics model such that the latent belief $q$ is informative for the policy.  However, for tasks with lower inherent physical ambiguity, the latent representation may be prone to mode collapse
if the downstream policy does not strictly demand an informative outcome embedding. To maintain an informative latent space in such cases, we incorporate an additional reconstruction loss to explicitly prevent representation collapse; see details in Appendix \ref{sec:reconstruction_loss}.

\subsection{Offline Data Generation}

For the model to effectively learn how to adapt from erroneous actions, the interaction histories seen during training must explicitly contain mispredicted actions. We therefore construct an offline meta-training dataset consisting of paired history sequences (for inner-loop adaptation) and query states with their ground-truth actions (for the outer-loop loss). To ensure a diverse distribution of outcomes, we collect these history sequences by executing a mix of random actions, which simulate typical test-time failures, and ground truth actions derived from privileged simulator knowledge.



\section{Experiment and Analysis}


We evaluate our proposed method against baselines  across multiple manipulation tasks. We follow the ISE evaluation task suite~\citep{ko2026implicitstateestimationvideo}, which includes the Slide Brick, Push Bar, Pick Bar, Open Box, and Turn Faucet, and also evaluate on an ambiguous door dataset~\citep{lihave,xiang2020sapien}. 
We also conduct a sim-to-real experiment of pick bar and push bar on a real-world uneven bar. In our instantiation of the method, we represent observations as point clouds, and actions are parameterized as a 6D vector 
that translates into task-specific physical interactions, such as defining a suction point and pull axis for opening a door, or specifying a pre-grasp waypoint for lifting an object.

\subsection{Quantitative Results}
\label{main:qualitative}


A key metric for online adaptation efficiency is the number of interaction attempts required for success. Beyond the benchmark baselines, we compare two variants sharing SCOUT's GMM policy head but replace test-time optimization: a \emph{raw GMM} without adaptation (Appendix~\ref{app:raw_gmm}) and a \emph{history-conditioned GMM} that encodes prior attempts $(o_t,a_t,o_{t+1})$ (Appendix~\ref{app:hist_gmm}). These isolate the benefits of interaction history and gradient-based belief updates, respectively.

\paragraph{Task Set 1: ISE Benchmark} Following ISE~\citep{ko2026implicitstateestimationvideo}, we evaluate on five tasks: Slide Brick, Push Bar, Pick Bar, Open Box, and Turn Faucet. All of the five tasks are highly ambiguous, so the policies must use trial and error to figure out the correct optimal action. For Slide Brick, Push Bar, and Pick Bar, the hidden physical property (friction for Slide Brick or center of mass for Push/Pick Bar) is continuous, which requires reasoning about the action outcome rather than only binary success or failure. The action outcome feedback can provide rich information to help update the policy. We compare against BC, CQL, the video-replanning baselines AVDC and ISE, and a PEARL-style probabilistic context baseline~\citep{rakelly2019pearl}; implementation details are provided in Appendices~\ref{app:ise_baselines} and~\ref{app:pearl}. As shown in Table~\ref{tab:adaptation_attempts}, SCOUT demonstrates the fastest adaptation across all five tasks compared to other methods. The improvement is especially significant when dealing with continuous hidden variables where more adaptation and reasoning ability is needed. History conditioning improves over the raw GMM on every task, but SCOUT needs least attempts thus adapts most efficiently. Statistical significance analysis can be found at Appendix~\ref{app:stats}.

\begin{table}[ht]
\centering
\vspace{-10pt}
\caption{\textbf{ISE benchmark results}. The average number of attempts (along with standard errors) needed for success across 400 rollouts. Lower numbers denote faster adaptation. Normalized score refers to an aggregated efficiency score that compares all methods against the performance of our method. The best performing method is bolded. SCOUT achieves top efficiency across all tasks.
}
\vspace{5pt}
\label{tab:adaptation_attempts}
\renewcommand{\arraystretch}{1.2} 
\resizebox{0.95\textwidth}{!}{
\begin{tabular}{l c c c c c c}
\toprule
\textbf{Method} & \textbf{Slide Brick} & \textbf{Push Bar} & \textbf{Pick Bar} & \textbf{Open Box} & \textbf{Turn Faucet} & \textbf{All (Normalized)}\\
\midrule
CQL~\citep{kumar2020conservative}  & 9.94 $\pm$ 0.18 & 10.96 $\pm$ 0.20 & 10.22 $\pm$ 0.19 & 4.32 $\pm$ 0.18 & 6.63 $\pm$ 0.21 & 3.17 $\pm$ 0.04\\
BC   & 10.24 $\pm$ 0.33 & 11.26 $\pm$ 0.30 & 9.79 $\pm$ 0.32 & 2.59 $\pm$  0.19 & 4.54 $\pm$  0.27 & 2.72 $\pm$ 0.05 \\
ADVC~\citep{advc2024} & 8.36 $\pm$ 0.27 & 7.83 $\pm$ 0.27 & 5.41 $\pm$ 0.22 & 2.82 $\pm$ 0.16 & 2.67 $\pm$ 0.16 & 1.91 $\pm$ 0.04 \\
ISE~\citep{ko2026implicitstateestimationvideo}  & 7.69 $\pm$ 0.26 & 5.26 $\pm$ 0.20 & 4.84 $\pm$ 0.18 & 2.25 $\pm$ 0.10 & 2.39 $\pm$ 0.14 & 1.61 $\pm$ 0.03\\
PEARL~\citep{rakelly2019pearl} & 5.03 $\pm$ 0.23 & 4.51 $\pm$ 0.20 & 4.89 $\pm$ 0.25 & 2.24 $\pm$ 0.12 & \textbf{1.54} $\pm$ 0.03 & 1.31 $\pm$ 0.03\\
\rowcolor{gray!12}Raw GMM(~\ref{app:raw_gmm})     & 6.39 $\pm$ 0.28 & 7.95 $\pm$ 0.25 & 4.96 $\pm$ 0.20 & 2.41 $\pm$ 0.10 & 2.22 $\pm$ 0.07 & 1.67 $\pm$ 0.03\\
\rowcolor{gray!12}History-cond. GMM (~\ref{app:hist_gmm})    & 3.92 $\pm$ 0.23 & 4.65 $\pm$ 0.22 & 3.36 $\pm$ 0.11 & 1.87 $\pm$ 0.05 & 1.61 $\pm$ 0.03 & 1.11 $\pm$ 0.02\\
\rowcolor{gray!25} Ours (SCOUT) & \textbf{3.37} $\pm$ 0.19 & \textbf{4.14} $\pm$ 0.18 & \textbf{2.83} $\pm$ 0.17 & \textbf{1.83} $\pm$ 0.05 & \textbf{1.54} $\pm$ 0.03 & \textbf{1.00} $\pm$ 0.02\\
\bottomrule
\end{tabular}%
}
\vspace{-5pt}
\end{table}

\paragraph{Task Set 2: Ambiguous Door Benchmark}

\begin{wraptable}{r}{0.7\textwidth} 
\centering
\caption{\textbf{Ambiguous door results}. SCOUT outperforms baselines without the high time cost required by generating multiple samples. (\# steps) for SCOUT indicates the number of adaptation steps used.}
\label{tab:door_results}
\renewcommand{\arraystretch}{1.3}
\resizebox{\linewidth}{!}{%
\begin{tabular}{l c >{\columncolor{gray!12}}c >{\columncolor{gray!12}}c c c >{\columncolor{gray!25}}c >{\columncolor{gray!25}}c}
\toprule
& \multicolumn{3}{c}{\textbf{Generator Only}} & \multicolumn{2}{c}{\textbf{Generator + Verifier}} & \multicolumn{2}{>{\columncolor{gray!25}}c}{\textbf{SCOUT}} \\
\cmidrule(lr){2-4} \cmidrule(lr){5-6} \cmidrule(lr){7-8}
& PNDiT~\citep{lihave} & \makecell[c]{Raw\\GMM} & \makecell[c]{History-cond. \\GMM} & \makecell[c]{HAVE~\citep{lihave} \\ (5 samples)} & \makecell[c]{HAVE~\citep{lihave} \\ (20 samples)} & \makecell[c]{SCOUT \\ (1 step)} & \makecell[c]{SCOUT \\ (5 steps)} \\
\midrule
Success Rate & 0.68 & 0.61 & 0.68  & 0.82 & 0.84 & 0.83 & \textbf{0.97} \\
Avg Steps    & 15.0 & 15.8 & 18.7 & 13.4 & 12.9 & 13.6 & \textbf{12.8} \\
Time (s)     & 0.463 & 0.002 & 0.004 & 0.757 & 1.876 & 0.012 & 0.052 \\
\bottomrule
\end{tabular}
}
\vspace{-10pt}
\end{wraptable}


We also test our method on the ambiguous door benchmark proposed in \citet{li2024flowbothd, lihave}, where doors with the same geometry can have different opening dynamics. The policy is expected to try different opening modes, and once it finds the correct mode, it should open the door all the way consistently. We report the success rate of fully opening the door within a maximum of 20 steps; we also report the average number of steps used for successful trials. From Table \ref{tab:door_results}, we can see that our method outperforms all of the baselines. The generator and verifier paradigm from \citet{lihave} has decent performance, but the time cost is much higher because this approach needs to generate multiple samples each time in order for the verifier to select the correct one, and the generator used in \citet{lihave} is a relatively slow diffusion model. The gap between history-conditioned GMM and SCOUT is much larger here (0.68 vs.\ 0.97), which we attribute to the longer histories (up to 20 steps) that the encoder must aggregate in one pass.

\begin{figure}[ht]
    \centering
    
    \begin{minipage}[c]{0.35\textwidth}
        \centering
        \vspace{-15pt}
        \resizebox{\linewidth}{!}{
            \renewcommand{\arraystretch}{1.3} 
            \begin{tabular}{l c c c}
                \toprule
                & \multicolumn{3}{c}{\textbf{CoM Position}} \\
                \cmidrule(lr){2-4}
                \textbf{Method} & \textbf{Left} & \textbf{Center} & \textbf{Right} \\
                \midrule
                \multicolumn{4}{l}{\textit{Pick Bar}} \\
                w/o TTT & $3.2 \pm 1.9$ & 3.6 $\pm$ 1.8 & $3.0 \pm 1.6$ \\
                \rowcolor{gray!15} SCOUT   & $\textbf{2.0} \pm 0.5$ & $\textbf{2.7} \pm 2.0$ & $\textbf{1.8} \pm 0.7$ \\
                \midrule
                \multicolumn{4}{l}{\textit{Push Bar}} \\
                w/o TTT & 6.1 $\pm$ 3.0 & 6.1 $\pm$ 2.5 & 4.6 $\pm$ 2.0 \\
                \rowcolor{gray!15} SCOUT   & $\textbf{2.9} \pm 0.8$ & $\textbf{3.7} \pm 0.5$ & $\textbf{2.1} \pm 2.0$ \\
                \bottomrule
            \end{tabular}
        }
    \end{minipage}
    \hfill
    \begin{minipage}[c]{0.6\textwidth}
        \centering
        \includegraphics[width=\textwidth]{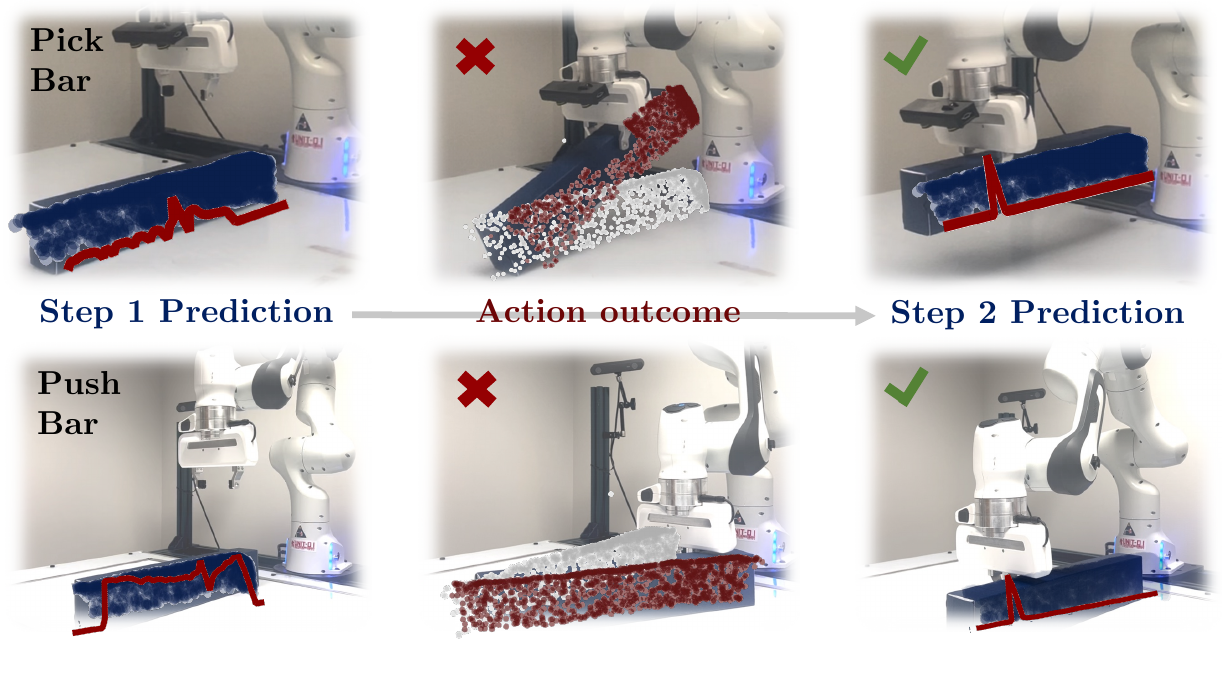}
    \end{minipage}

    \vspace{-0.2cm} 
    \caption{\textbf{Real world uneven bar pickup and push results}. Left: The number of attempts to success with and without test-time training, reported with 95\% confidence interval. Right:  A visualization of the policy prediction weights across the x-axis of the uneven bar before and after adaptation. We can see that SCOUT efficiently adapts to different Center of Mass positions. 
    }
    \label{fig:realworld}
    \vspace{-10pt}
\end{figure}

\paragraph{Real World Task: Uneven Bar Pickup \& Push} We evaluate our method, trained exclusively in simulation, on a physical uneven bar to assess its sim-to-real transfer performance. We collect 10 rollouts for each method and each Center of Mass (CoM) position, with a maximum of 10 steps per rollout. As shown in Fig. \ref{fig:realworld}
, our method can effectively reason about the real world action outcomes and efficiently adapt the belief space to make better action predictions, and requires significantly fewer attempts than the non-TTT baseline on both tasks (Appendix~\ref{app:stats}). 


\subsection{Analysis}

\textbf{Belief Latent Space Visualization.} To verify that the adapted belief $q_k$ encodes meaningful object dynamics, we visualize its evolution during test-time adaptation. In our door opening task, closed doors present a challenging four-mode visual ambiguity (pull/push, left/right). If the latent vector $q_k$ genuinely captures underlying physical dynamics, successive adaptation updates should progressively move the uninformative, shared prior $q_0$ toward the correct kinematic mode. 

\begin{wrapfigure}{r}{0.6\textwidth}
    \centering
    \vspace{-8pt}
    \includegraphics[width=\linewidth]{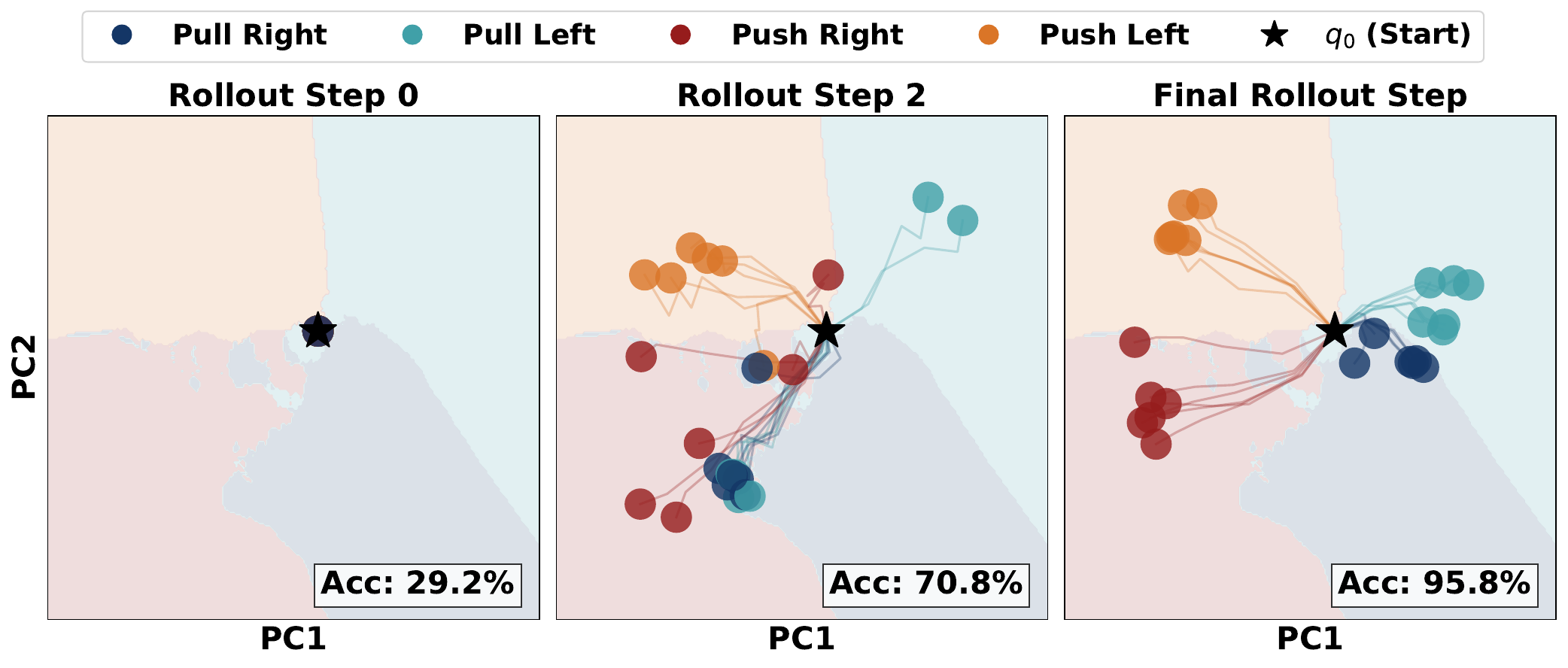}
    \caption{\textbf{Belief Latent Space Evolution.} PCA projection of the belief latent $q_k$ across three stages of adaptation. 
    Points are colored by their ground-truth motion mode, while background regions represent kNN-estimated decision boundaries. As test-time training progresses, the latents successfully migrate from the shared prior $q_0$ into distinct, well-separated kinematic clusters.}
    \label{fig:pca}
    \vspace{-5pt}
\end{wrapfigure}

We project the adapted $q_k$ vectors onto their top two principal components (PC), coloring each point according to its ground-truth mode. As illustrated in Figure~\ref{fig:pca}, the background regions denote latent decision boundaries estimated via a $k$-nearest neighbor (kNN) classifier. By tracking the spatial distribution across the three rollout stages, we observe a clear progression: at Step 0, all latents originate at the shared prior $q_0$; by Step 2, intermediate clusters begin to form as the model integrates early interaction history. By the Final step, the latents have fully migrated into well-separated regions corresponding to their true kinematics. This visual evidence of progressive convergence confirms that our method effectively maps accumulated observation history into a dynamically meaningful belief state.

\begin{figure}[t]
    \centering
    \includegraphics[width=1.0\linewidth]{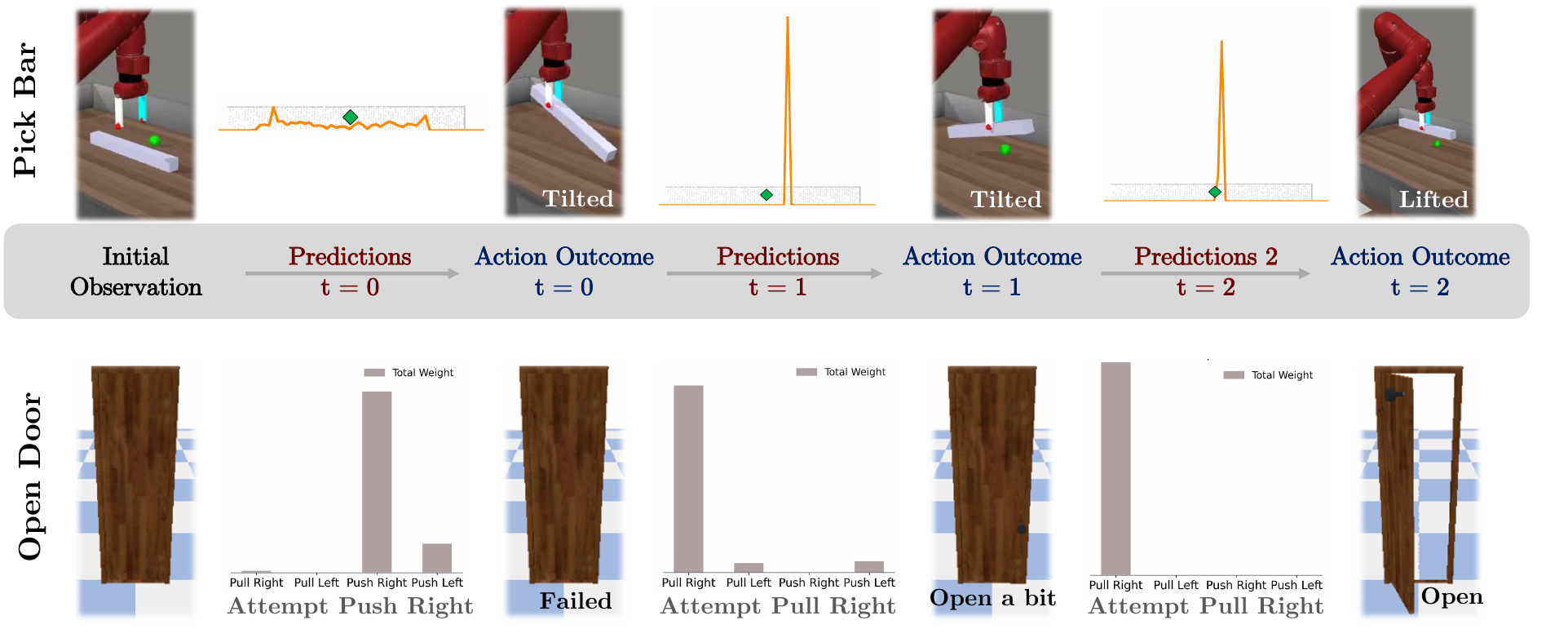}
    \caption{
    \textbf{Evolution of predicted action distributions across rollout steps}. (Top) Pick Bar: Despite high initial variance, the predicted grasp distribution quickly converges toward the optimal center of mass. (Bottom) Open Door: Probability mass across discrete action modes successfully concentrates onto the correct target action (i.e., Pull Right) as online adaptation progresses.
    }
    \label{fig:door_dist}
    \vspace{-5pt}
\end{figure}

\textbf{Adaptation Visualization.} To characterize how test-time adaptation shapes action predictions over a rollout, we visualize the evolution of action weight distributions in continuous (Pick Bar) and discrete (Open Door) tasks (Fig.~\ref{fig:door_dist}). For the continuous task, each column displays the scene observation alongside predicted weight assignments per step, marking the ground-truth center of mass with a green diamond. For the discrete task, we categorize the continuous action predictions into the four action modes, and calculate the total GMM weight assigned to each. Initially, predictions reflect either a highly uncertain prior (Pick Bar) or an incorrect initial bias (Open Door). As interaction history accumulates through failed attempts, adaptation rapidly corrects these initial beliefs, concentrating the probability mass onto the optimal actions, converging toward the true center of mass or isolating the correct articulation direction. This confirms that test-time updates effectively shift the belief from a generalized or biased prior to a localized, target-specific representation.

\vspace{-12pt}

\paragraph{Generalization to OOD Dynamics}

\begin{wraptable}{r}{0.55\linewidth}
\vspace{-12pt}
\centering
\caption{\textbf{Held-out dynamics ranges.} Average attempts to success when training on parameter segments R1/R3/R5 and evaluating on all five. The last column is the raw GMM trained on the \emph{full} range (Table~\ref{tab:adaptation_attempts}).}
\label{tab:ood}
\renewcommand{\arraystretch}{1.15}
\resizebox{\linewidth}{!}{%
\begin{tabular}{lcccc>{\columncolor{gray!7}}c}
\toprule
& \multicolumn{2}{c}{Trained} & \multicolumn{2}{c}{Held-out} & \multicolumn{1}{c}{Raw GMM} \\
\cmidrule(lr){2-3}\cmidrule(lr){4-5}
Task & R1/R3/R5 & Avg & R2/R4 & Avg & \multicolumn{1}{c}{(full range)} \\
\midrule
Pick Bar    & 2.8/2.2/1.7 & 2.25 & 4.0/5.1 & 4.56 & 4.96 \\
Push Bar    & 3.7/5.7/3.9 & 4.48 & 5.9/6.8 & 6.34 & 7.95 \\
Slide Brick & 4.6/3.0/2.1 & 3.14 & 3.0/3.7 & 3.36 & 6.39 \\
\bottomrule
\end{tabular}
}
\vspace{-8pt}
\end{wraptable}
To test whether adaptation generalizes to physical parameters never seen during meta-training, we divide each hidden-parameter range (center of mass for Pick Bar and Push Bar, and friction for Slide Brick) into five ordered segments, R1--R5. We train on R1/R3/R5 and evaluate on the held-out R2/R4. In contrast, the raw GMM is trained on the full range and therefore sees these values during training. As shown in Table~\ref{tab:ood}, held-out performance on Slide Brick (3.36) is nearly identical to performance on the training segments (3.14). On the two center-of-mass tasks (Pick Bar and Push Bar), the held-out settings require more attempts, yet SCOUT still adapts faster than this non-adapting baseline. These results show that the inner loop updates the belief in a useful direction for unseen dynamics, though less efficiently for interpolated values than for those observed during training.



\section{Conclusion}

In this paper, we introduce SCOUT, a dynamics-aware meta-learning framework that enables robotic manipulation policies to rapidly adapt based on the gap between the believed dynamics and the actual action outcomes. Unlike existing methods that rely on sparse rewards, 
SCOUT leverages rich action-outcome mismatches by coupling an action-prediction policy with a forward dynamics model via a shared belief latent space. Through meta-learning, 
the agent learns to update this latent belief at deployment  based on dynamics mispredictions. This enables the policy to adapt at test-time without catastrophic forgetting of the pre-trained policy weights. 
Our evaluations demonstrate that SCOUT effectively adapts to different dynamics and significantly accelerates the adaptation efficiency across different simulated benchmarks and in the real world.
\section{Limitations}


A limitation of the current work is that the current experimental validation is restricted to physical ambiguities, leaving broader classes of manipulation failures for future work. While the shared-belief formulation is general in principle, the present framework has not yet been evaluated on failures arising from complex geometric constraints, perception errors, or other sources of uncertainties in robotics. In addition, the current implementation has not been integrated with large-scale foundation policy architectures. Extending our 
adaptation 
mechanism to 
diffusion-based generative policies and Vision-Language-Action (VLA) models remains an important direction for future work.



\clearpage


\acknowledgments{This material is based upon work supported by ONR MURI N00014-24-1-2748. We thank Kallol Saha for valuable discussions about paper writing, Kyutae Sim and Mino Nakura Fan for helping with the real robot setup.}


\bibliography{main}  

\clearpage
\newpage
\appendix
\addcontentsline{toc}{section}{Appendix} 
\part{Appendix} 
\parttoc 

\section{Method Details}

\subsection{Action and Action Outcome Representation}
\label{sec:representation}

\paragraph{Action} The action representation differs across task families. In all ISE tasks, $a_t \in \mathbb{R}^3$. For Push Bar and Pick Bar, this 3D vector specifies the contact point where the Sawyer hand pushes or picks the bar. For Slide Brick, it specifies the highest point to which the Sawyer hand slides the brick before the brick moves freely downward. For Turn Faucet and Open Box, which each have two discrete options, the 3D vector specifies the final target position reached by the Sawyer hand. In the ambiguous door task, $a_t \in \mathbb{R}^6$ and is parameterized as $(x, y, z, \Delta x, \Delta y, \Delta z)$: the first three dimensions specify the point where the suction gripper attaches to the door, and the last three dimensions specify the directional vector for the movement.

\paragraph{Action Outcome} As described in the main method, the agent executes an action $a_t$ from observation $o_t$ and receives a post-action outcome $o_{t+1}$. For the ISE benchmark tasks, we represent this outcome directly as the point cloud observation after the action. For the ambiguous door task, we represent the outcome as a per-point flow field, i.e., the displacement vector from each point in the pre-action point cloud to its corresponding point after interaction.

\subsection{Gaussian Mixture Model (GMM)}
\label{sec:policy_network}
The policy network is a GMM head conditioned on the belief latent $q$ and the current point cloud observation $o_t$. 

Each point is first embedded via a linear projection into a $d_{\text{model}}$-dimensional feature space, then processed by $L$ FiLM-conditioned transformer blocks, where in each block $q$ is linearly projected to a per-channel scale $\gamma$ and shift $\beta$ that are applied to the feed-forward output.The per-point features are then concatenated with a broadcast copy of $q$ and passed through two MLP heads to produce mixture weights $w \in \Delta^{K-1}$ and component means $\{\mu_k\}_{k=1}^{K}$, where each mean consists of a contact point offset relative to the corresponding point cloud location and a unit-normalized push direction. In addition to FiLM conditioning within each transformer block, $q$ is also fused at the output via zero-initialized residual projections $q \mapsto \Delta w$ and $q \mapsto \Delta \mu$, which are added directly to the predicted mixture weights and means. Zero initialization ensures that at the start of training the GMM output is determined purely by the point cloud geometry, with $q$'s direct influence 
growing as training proceeds.

In our setting, we use $d_{\text{model}} = 128$, $L = 4$ transformer layers, and 4 attention heads. We set the number of GMM components depends on the task ambiguity: for the ISE tasks with continuous hidden physical properties (Slide Brick, Push Bar, and Pick Bar), we use $K = 4$ components; for the binary ISE tasks (Open Box and Turn Faucet), we use $K = 2$ components. The ambiguous door benchmark also uses $K = 4$ components to capture its four possible opening modes.

\subsection{Outcome Encoder}
\label{sec:outcome_encoder}
The Outcome Encoder $E$ takes as input the action outcome. For ISE tasks, this is the post-action point cloud $o_{t+1} \in \mathbb{R}^{N \times 3}$. For the ambiguous door task, we use the observed flow $f \in \mathbb{R}^{N \times 3}$ (the per-point flow field induced by the action) together with the post-action point cloud. Accordingly, we have different outcome encoder model architectures for ISE and ambiguous door tasks. 


\textbf{ISE tasks.} The outcome encoder is a transformer over the post-action point cloud. It first embeds each point in $o_{t+1}$ with a linear projection into a $d_{\text{model}}$-dimensional feature space, then applies a stack of self-attention blocks. The resulting per-point features are normalized, mean-pooled across points, and projected through a linear layer followed by layer normalization to produce $z_{\text{target}, t} = E(o_{t+1}) \in \mathbb{R}^{d_z}$. The final layer normalization helps avoid degenerate constant outputs while still using only the observed post-action point cloud.

\textbf{Ambiguous door task.} The encoder uses both the post-action geometry and the observed flow. It embeds $o_{t+1}$ and $f$ separately via two linear projections into a shared $d_{\text{model}}$-dimensional feature space, yielding a position embedding from $o_{t+1}$ and a flow embedding from $f$. In the first attention block, queries and keys receive the sum of both embeddings, while values receive the flow embedding alone. Subsequent blocks apply standard self-attention over the resulting feature stream. The final per-point features are mean-pooled and projected through a linear layer followed by layer normalization to produce the outcome latent $z_{\text{target}, t}$.

In our setting, we use $N = 1200$, $d_z = 64$, $d_{\text{model}} = 128$, 2 attention layers, and 4 attention heads.

\subsection{Door-Specific Reconstruction Loss and Flow Decoder}
\label{sec:reconstruction_loss}
In highly ambiguous environments (e.g., the ISE tasks), the inner-loop adaptation is essential for solving the task. Because the policy relies heavily on the adapted belief, the outer-loop meta-objective backpropagates strong gradients through the inner-loop updates, actively forcing the outcome encoder to learn an informative latent space. However, in the ambiguous door experiments, certain states (such as a partially opened door) lack physical ambiguity. With these instances occupying an inevitable part of the dataset, the model learns a shortcut: the policy can minimize the outer-loop loss by relying solely on the query observation, effectively ignoring the adapted belief. When the policy ignores this belief, the gradients flowing from the outer loop to the outcome encoder vanish. Without any gradient supervision to maintain a meaningful representation, the latent space is prone to mode collapse, degrading into a constant embedding that zeros out the inner-loop gradients and renders the adaptation mechanism entirely ineffective. 

To prevent the representation collapse of the Outcome Encoder $E$, we introduce an auxiliary Flow Decoder $\text{Dec}$ that reconstructs the observed flow $o_{\text{t+1}}$ from the post-action point 
cloud and the outcome latent.
If $z_{\text{target}}$ collapses to a constant, the decoder cannot reconstruct 
the per-sample flow and the reconstruction loss rises, providing direct gradient 
pressure on $E$ to retain informative latent representations.

The Flow Decoder is a point-wise MLP: each point $p_i \in \mathbb{R}^3$ is 
concatenated with the global latent $z_{\text{target}}$ and passed through two 
hidden layers with GELU activations to predict the flow $o_{t+1}$ at that point.
The reconstruction loss is:
\begin{equation*}
    L_{\text{recon}} = \bigl\|\text{Dec}(E(o_{\text{t+1}})) - o_{\text{t+1}}\bigr\|_2^2,
\end{equation*}
and is aggregated over both the query step and all history steps to provide 
denser supervision.
This loss is computed exclusively in the outer loop and does not enter the 
inner-loop adaptation.

In our setting, the Flow Decoder uses a hidden dimension of 128 and 
$\lambda_{\text{recon}} = 1.0$.

\section{Baseline and Ablation Details}

\subsection{ISE Benchmark Baselines}
\label{app:ise_baselines}

We follow the baseline definitions from the ISE benchmark~\citep{ko2026implicitstateestimationvideo}. \textbf{BC} is a state-embedding-conditioned behavior-cloning policy trained with privileged ground-truth action labels; both the current state and a DINOv2 embedding of the physical configuration condition the policy. \textbf{CQL}~\citep{kumar2020conservative} uses the same state embedding in an offline actor--critic formulation: the critic receives the state, action, and embedding, while the policy receives the state and embedding. These two methods provide imitation-learning and offline-RL references for policies that are given an explicit representation of the hidden configuration.

\textbf{AVDC}~\citep{avdc2024} is a video-planning baseline that generates an actionless future video and converts it into robot motion using dense visual correspondences; it replans after failure but does not update an internal belief from interaction history. \textbf{ISE} augments this pipeline with interaction-time adaptation. It encodes interaction videos into a latent state embedding, retrieves and refines an embedding using observed failed interactions, and conditions subsequent video-plan generation on that embedding. It also maintains failed-interaction and failed-plan buffers and selects candidate plans that differ from previous failures. Thus, ISE performs online adaptation through video replanning and implicit state estimation, whereas SCOUT directly adapts the belief latent of a point-cloud action policy using action-outcome prediction error.

\subsection{PEARL-Style Probabilistic Context Baseline}
\label{app:pearl}

PEARL~\citep{rakelly2019pearl} performs rapid adaptation by inferring a probabilistic latent task variable from accumulated experience and conditioning the policy on samples from that posterior. We adapt this context-inference mechanism to our attempt-level imitation-learning setting. Each previous attempt provides a context tuple $c_i=(P_i^{\mathrm{before}},a_i,P_i^{\mathrm{after}})$ containing the point cloud before execution, the selected 3D target, and the observed post-action point cloud. Separate two-layer self-attention encoders process the before and after point clouds; the before branch additionally receives each point's displacement from the attempted target. Their mean-pooled features are fused to predict the mean $\mu_i$ and diagonal variance $v_i$ of a Gaussian factor over a 64-dimensional latent $z$. Following PEARL, the factors are combined with a permutation-invariant product of Gaussians:
\begin{equation}
    v^{-1}=\sum_{i=1}^{H}v_i^{-1},
    \qquad
    \mu=v\sum_{i=1}^{H}\mu_i v_i^{-1}.
\end{equation}
An empty history uses the fixed prior $\mathcal{N}(0,I)$. Each point cloud is sampled to 1,200 points, and the current observation, history, and actions share the coordinate normalization defined by the current point cloud.

The sampled latent and current point cloud condition a GMM action head with the same task-dependent number of components used by SCOUT. We jointly train the context encoder and action head to minimize the negative log likelihood of the expert action plus $0.01D_{\mathrm{KL}}(q_\phi(z\mid c_{1:H})\,\|\,\mathcal{N}(0,I))$, using the reparameterization trick. On 20\% of training examples, history is omitted and $z=0$ to train the first-attempt predictor. At evaluation, one posterior sample is held fixed while the selected target is executed; after a failure, the resulting attempt is appended to the context before the next posterior update. The encoder receives the observed post-attempt point cloud but no reward or explicit success flag. This is a PEARL-style baseline rather than a reproduction of the full actor--critic algorithm: it preserves probabilistic context inference, the information bottleneck, and posterior sampling, while replacing SAC and critic learning with supervised GMM action prediction for fair comparison with our method.

\subsection{Raw GMM}
\label{app:raw_gmm}

The raw GMM baseline (``Raw GMM'' in Tables~\ref{tab:adaptation_attempts} and~\ref{tab:door_results}) keeps only the policy component of SCOUT: the same point-cloud based GMM head (Appendix~\ref{sec:policy_network}) without the belief latent $q$, the dynamics model, or the outcome encoder. It is trained independently on the same demonstration data with the standard imitation objective, with no inner loop and no meta-learning, and at test time it simply samples from its predicted mixture at every attempt. We use an independently trained policy rather than SCOUT with $q$ frozen at $q_0$ so that the baseline's inputs remain in-distribution. This is also the ``w/o TTT'' baseline in the real-world experiments (Fig.\ref{fig:realworld}). The comparison therefore measures what the full adaptation machinery adds on top of the policy architecture. As Table~\ref{tab:adaptation_attempts} shows, the raw GMM already provides a reasonable action prior on the ISE tasks, but requires $1.67\times$ as many attempts as SCOUT overall, showing that the GMM head alone isn't capable of online adaptation.

\subsection{History-conditioned GMM}
\label{app:hist_gmm}
To test whether test-time optimization is necessary, or whether the same information could be extracted by a learned encoder in a single forward pass, we evaluate a history-conditioned GMM policy that directly takes in previous interaction attempts without performing test-time optimization. The architecture has two components:

\textbf{GMM head.} We use the same GMM head architecture as in the main method: a FiLM-conditioned transformer processes the current point cloud observation, and the resulting point features parameterize a Gaussian mixture distribution over actions. The only difference is the conditioning signal. Instead of using the test-time-adapted belief latent $q$, this baseline conditions the GMM head on a feed-forward history context vector produced by the history encoder below.

\textbf{History encoder.} The history encoder maps each prior attempt $(o_t, a_t, o_{t+1})$ into a per-attempt token: a before-branch processes the pre-action point cloud concatenated with the action-conditioned point features through a self-attention trunk and mean pooling, while an after-branch processes the post-action point cloud with a separate self-attention trunk and mean pooling. These two summaries are fused with an MLP to form an attempt token. A cross-attempt self-attention module then mixes the variable-length set of attempt tokens, mean-pools the result, and projects it to a history context vector in $\mathbb{R}^{d_z}$. For the first attempt, when no history is available, we use a learned null context.



\subsection{Reconstruction Loss}
\label{app:ablation}

We ablate the reconstruction loss we added for the ambiguous door benchmark. In \textbf{w/o reconstruction loss}, the \texttt{FlowDecoder} branch and its objective $\mathcal{L}_\text{recon}$ (Eq.~1) are removed from the outer-loop objective by setting $\lambda_{\text{recon}} = 0$. Results are reported in Table~\ref{tab:ablation} under both the single-step ($k{=}1$) and five-step ($k{=}5$) adaptation settings.

While removing the reconstruction loss still yields improvement over the raw GMM (0.71 vs.\ 0.61 success), it shows suboptimal adaptation behavior due to representation degradation. This validates our motivation for adding the reconstruction head specifically to prevent the \texttt{OutcomeEncoder} from collapsing to a near-constant latent $z$, a failure mode that is otherwise viable on low-ambiguity tasks since nothing in the inner-loop latent MSE alone penalizes a constant encoder. Without $\mathcal{L}_\text{recon}$, the encoder has no direct incentive to retain outcome-specific information in $z$, degrading the inner loop's ability to adapt $q$ meaningfully from history, and in turn degrading the quality of GMM proposals at test time. This term is not always required: on the fully ambiguous ISE tasks the outer loss alone prevents collapse, and we use it only for the ambiguous door benchmark.

\begin{figure}[t]
  \centering
  \includegraphics[width=0.8\linewidth]{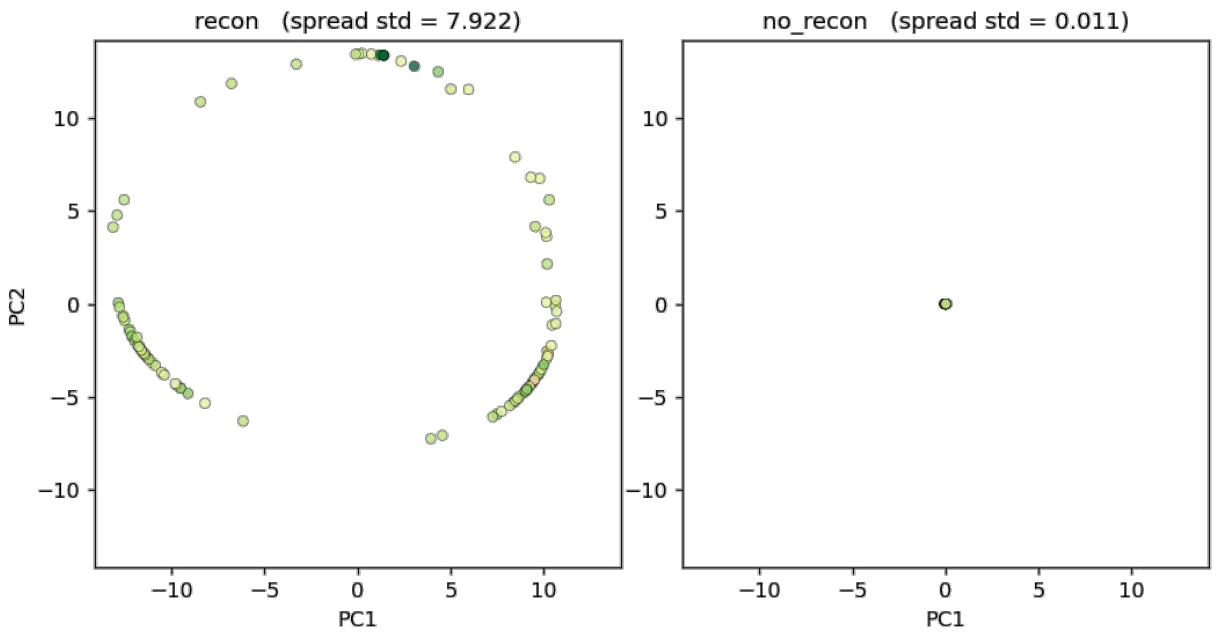}
  \caption{%
    \textbf{PCA projection of OutcomeEncoder latents $z$ on 120 validation outcomes}, with (left) and without (right) the reconstruction loss, drawn on shared axes. Because the encoder's projection head ends in a \texttt{LayerNorm}, every $z$ has a near-constant norm ($\lVert z \rVert \approx 13$--$20$), so collapse manifests as a common \emph{direction} rather than a vanishing magnitude. With $\mathcal{L}_\text{recon}$ the latents span a wide arc (per-axis spread std $7.9$), indicating that outcome-specific information is retained in $z$. Without it the latents collapse to a single point at the origin (spread std $0.01$), a $644\times$ reduction in mean pairwise distance relative to the reconstruction model.}
  \label{fig:recon-collapse-pca}
\end{figure}

We visualize this collapse directly in Fig.~\ref{fig:recon-collapse-pca}. We encode $120$ validation outcomes with each \texttt{OutcomeEncoder} and compare the resulting latents $z$. With the reconstruction head, distinct outcomes are mapped to well-separated latents; without it, the encoder maps every outcome to an almost identical vector. Since the proposal head is conditioned on $q$, which the inner loop can only move by matching this constant target, a collapsed $z$ leaves the GMM proposals near their unadapted state---explaining the suboptimal adaptation observed above.

\begin{table}[h]
  \centering
  \caption{%
    \textbf{Reconstruction loss ablation on the ambiguous doors}.
    Success Rate ($\uparrow$) and Avg.\ Steps ($\downarrow$) for SCOUT with and without the reconstruction loss, each evaluated with 1 and 5 adaptation steps.
    \textbf{Bold} denotes the best result per metric.
  }
  \vspace{10pt}
  \label{tab:ablation}
  \setlength{\tabcolsep}{8pt}
  \renewcommand{\arraystretch}{1.15}
  \begin{tabular}{l cc >{\columncolor{gray!15}}c >{\columncolor{gray!15}}c}
    \toprule
    & \multicolumn{2}{c}{\textbf{SCOUT (w/o Recon. loss)}}
    & \multicolumn{2}{>{\columncolor{gray!15}}c}{\textbf{SCOUT (Full)}} \\
    \cmidrule(lr){2-3}\cmidrule(lr){4-5}
    & 1 step & 5 steps & 1 step & 5 steps \\
    \midrule
    Success Rate  & 0.71 & 0.73 & 0.83 & \textbf{0.97} \\
    Avg Steps     & 14.2 & 14.7 & 13.6 & \textbf{12.8} \\
    \bottomrule
  \end{tabular}
\end{table}

\section{Statistical Significance}
\label{app:stats}

\subsection{Simulation: SCOUT vs.\ History-conditioned GMM}

Every ISE entry in Table~\ref{tab:adaptation_attempts} is the mean number of attempts to success over 400 rollouts, reported with its standard error. The history-conditioned GMM (Appendix~\ref{app:hist_gmm}) is the closest encoder-based alternative to SCOUT, since it shares the same policy head and sees the same interaction history but infers its conditioning vector in a single forward pass rather than by gradient updates. To test whether SCOUT's advantage over it is significant, we compare the per-rollout attempt counts of the two methods on each task with a two-sided Mann--Whitney $U$ test ($n = 400$ per method), with Holm correction for multiple comparisons.

SCOUT requires substantially fewer attempts than the history-conditioned GMM on all three continuous-parameter tasks: 3.37 versus 3.92 on Slide Brick ($p_{\mathrm{adj}} = 4.93\times10^{-24}$), 4.14 versus 4.65 on Push Bar ($p_{\mathrm{adj}} = 7.29\times10^{-13}$), and 2.83 versus 3.36 on Pick Bar ($p_{\mathrm{adj}} = 3.75\times10^{-26}$). Thus, the advantage is statistically significant on all three continuous-parameter tasks ($p_{\mathrm{adj}} < 10^{-12}$ on each), where the geometry of the action outcome carries information about the hidden parameter that a learned encoder must capture implicitly but the inner loop corrects explicitly. On the two binary tasks, SCOUT also uses fewer attempts on average: 1.83 versus 1.87 on Open Box ($p_{\mathrm{adj}} = 7.40\times10^{-5}$) and 1.54 versus 1.61 on Turn Faucet ($p_{\mathrm{adj}} = 1$). In these tasks, a single failed attempt largely reveals the correct mode, leaving less room for gradient-based refinement to improve performance.


\subsection{Real World: SCOUT vs.\ Non-TTT Baseline}

\textbf{Design.} The real-world experiment has 120 trials in total: 2 tasks (Pick Bar, Push Bar) $\times$ 3 center-of-mass positions (left, middle, right) $\times$ 2 methods (SCOUT and the non-TTT baseline of Appendix~\ref{app:raw_gmm}) $\times$ 10 rollouts. Each rollout is an independent episode with a fresh reset, and the two methods are evaluated on separate rollouts. The response variable is the number of attempts to success, an integer capped at the 10-step limit; a rollout that does not succeed within 10 steps is recorded as 10.

\textbf{Test.} The confidence intervals in Fig.~\ref{fig:realworld} are wide because the attempt counts are small bounded integers with many ties and a long right tail from the capped failures; the data are neither continuous nor approximately normal. We therefore use the two-sided Mann--Whitney $U$ test, which compares the two methods by rank rather than by mean and is unaffected by the cap. Because ties are frequent, we use the asymptotic normal approximation with tie and continuity corrections rather than the exact distribution, which assumes no ties. With only 10 rollouts per cell, we treat the comparison pooled over the three center-of-mass positions within each task ($n = 30$ vs.\ 30) as the primary analysis. Pooling is fair here because the design is balanced: every position contributes 10 rollouts to each method, so position cannot favor either side. We apply a Holm--Bonferroni correction across the two primary tests.

\begin{table}[h]
  \centering
  \caption{\textbf{Real-world significance tests, pooled over CoM positions.} Two-sided Mann--Whitney $U$ with tie and continuity correction, $n=30$ rollouts per method. $P_{\text{sup}}$ is the probability that a random SCOUT rollout needs fewer attempts than a random baseline rollout.}
  \label{tab:stats_pooled}
  \renewcommand{\arraystretch}{1.15}
  \resizebox{\textwidth}{!}{%
  \begin{tabular}{l cc cc c cc c}
    \toprule
    & \multicolumn{2}{c}{Median} & \multicolumn{2}{c}{Mean} & & \multicolumn{2}{c}{$p$} & \\
    \cmidrule(lr){2-3}\cmidrule(lr){4-5}\cmidrule(lr){7-8}
    Task & SCOUT & Baseline & SCOUT & Baseline & $U$ & raw & Holm & $P_{\text{sup}}$ \\
    \midrule
    Pick Bar & 2 & 3 & 2.17 & 3.27 & 311.5 & 0.035 & 0.035 & 0.65 \\
    Push Bar & 3 & 4 & 2.90 & 5.60 & 250.0 & 0.003 & 0.005 & 0.72 \\
    \bottomrule
  \end{tabular}%
  }
\end{table}

\textbf{Effect size.} Beyond the $p$-value, the $U$ statistic has a direct interpretation: over all $30 \times 30$ pairs of one SCOUT rollout and one baseline rollout, $U$ counts the pairs in which SCOUT needed fewer attempts (ties count one half). Dividing by the number of pairs gives $P_{\text{sup}}$, the probability that a randomly chosen SCOUT rollout beats a randomly chosen baseline rollout; $0.5$ means no difference.

\textbf{Results.} SCOUT needs significantly fewer attempts on both tasks (Table~\ref{tab:stats_pooled}): Pick Bar $p = 0.035$ (Holm-adjusted $0.035$) and Push Bar $p = 0.003$ (Holm-adjusted $0.005$). A random SCOUT rollout beats a random baseline rollout $65\%$ of the time on Pick Bar and $72\%$ of the time on Push Bar.

\section{Additional Analysis}
\label{app:analysis}

\subsection{Robustness to Estimated Flow}
\label{app:est_flow}

The main door experiments use the simulator's ground-truth per-point flow as the action outcome (Appendix~\ref{sec:representation}). On a real robot this flow must be estimated from observations, so we test whether SCOUT's adaptation survives the noise of an estimated outcome. At test time we replace the ground-truth flow with flow estimated by point tracking system~\citep{ngo2025delta} on the pre/post point clouds.

\begin{table}[h]
  \centering
  \caption{%
    \textbf{Robustness to estimated flow on the ambiguous doors}.
    Success Rate ($\uparrow$) and Avg.\ Steps ($\downarrow$) for SCOUT when the action outcome is the simulator's ground-truth flow (as in Table~\ref{tab:door_results}) or flow estimated at test time.
  }
  \vspace{10pt}
  \label{tab:est_flow}
  \setlength{\tabcolsep}{8pt}
  \renewcommand{\arraystretch}{1.15}
  \begin{tabular}{l cc cc}
    \toprule
    & \multicolumn{2}{c}{\textbf{Estimated flow}}
    & \multicolumn{2}{c}{\textbf{Ground-truth flow}} \\
    \cmidrule(lr){2-3}\cmidrule(lr){4-5}
    & 1 step & 5 steps & 1 step & 5 steps \\
    \midrule
    Success Rate  & 0.83  & 0.96  & 0.83 & \textbf{0.97} \\
    Avg Steps     & 15.86 & 14.94 & 13.6 & \textbf{12.8} \\
    \bottomrule
  \end{tabular}
\end{table}

As shown in Table~\ref{tab:est_flow}, the success rate is essentially unchanged: 0.83 with one adaptation step and 0.96 with five, versus 0.83 and 0.97 with ground-truth flow. Successful trials take somewhat longer with estimated flow (15.9 / 14.9 vs.\ 13.6 / 12.8 steps), suggesting that a noisier outcome slows the convergence of the belief without preventing it. The inner-loop update therefore does not depend on privileged simulator information at test time.

\subsection{What Does the Belief Latent Encode?}
\label{app:belief_probe}

The main text shows qualitatively that the adapted belief $q_k$ separates the four door modes (Fig.~\ref{fig:pca}). Here we ask more directly whether $q$ encodes the hidden physical variable itself, in three ways: by transferring a belief between objects that share physics, by regressing the physical parameter from $q$, and by decoding $q$ during real-world adaptation.

\paragraph{Cross-trajectory belief transfer.}
If $q_k$ captures the physical property of the door rather than memorizing the particular interaction history, it should transfer between doors that share dynamics but differ in geometry. 
We test this directly on the ambiguous-door
benchmark: we run the inner loop ($5$ steps) on door $A$'s three-failure history and use the resulting $q_k$ to condition the policy on a \emph{different} door $B$ that shares $A$'s articulation mode but differs in appearance, without any further adaptation on $B$. Door $B$ is reset to its closed pose where the correct action is the most ambiguous, and we evaluate a single step on that observation. This yields $236$ ordered cross-mesh pairs.
We report the negative log-likelihood of the expert action on B under the policy's GMM, and \emph{top-$100$ mode accuracy}: the fraction of the policy's $100$ highest-weight mixture components whose predicted direction matches the ground-truth flow of $B$'s true mode rather than one of the other three (chance $0.25$).

\begin{table}[h]
  \centering
  \caption{%
    \textbf{Cross-trajectory belief transfer on the ambiguous doors}.
    The policy is evaluated on query door B using a belief adapted on B itself, on a different door A with the same articulation mode, or not adapted at all.
  }
  \vspace{10pt}
  \label{tab:cross_traj}
  \setlength{\tabcolsep}{8pt}
  \renewcommand{\arraystretch}{1.15}
  \begin{tabular}{l c c}
    \toprule
    \textbf{Belief used on door B} & \textbf{NLL} ($\downarrow$) & \textbf{Top-100 acc.} ($\uparrow$) \\
    \midrule
    No adaptation ($q = q_0$)                 & 6.41 & 0.26 \\
    Adapted on B itself                       & 3.62 & 0.89 \\
    Adapted on a different door A (same mode) & \textbf{3.59} & \textbf{0.91} \\
    \bottomrule
  \end{tabular}
\end{table}

As shown in Table~\ref{tab:cross_traj}, a belief adapted on a different door performs as well as one adapted on the query door itself (NLL 3.59 vs.\ 3.62; top-100 accuracy 0.91 vs.\ 0.89), and both far outperform the unadapted prior (6.41; 0.26). Transfer under matched physics but mismatched geometry indicates that $q$ encodes the physical variable, not trajectory-specific details.


\paragraph{Decoding the physical parameter from $q$.}
The previous qualitative analysis of the latent focuses on doors, where the hidden latent represents discrete modes. To further test whether $q$ also encodes a \emph{continuous} hidden parameter, we collect the adapted belief $q_k$ at the successful action in Pick Bar and Push Bar simulation trajectories and train an MLP probe to regress the ground-truth bar-relative center of mass from $q_k$ alone, with all SCOUT weights frozen. We roll out 800 trajectories per task and exclude failed trajectories and trajectories that succeed on the first attempt, before $q$ is adapted. This leaves 621 Pick Bar and 630 Push Bar trajectories, with one probe sample per trajectory. The probe is an MLP with two hidden layers of 128 and 64 units with ReLU activations after each hidden layer; we report $R^2$ from five-fold cross-validation on held-out trajectories. It reaches $R^2 = 0.982$ on Pick Bar and $R^2 = 0.965$ on Push Bar. Although $q$ is never supervised with the physical parameter, it accurately predicts that parameter at the successful action.

\begin{figure}[h]
  \centering
  \includegraphics[width=0.85\linewidth]{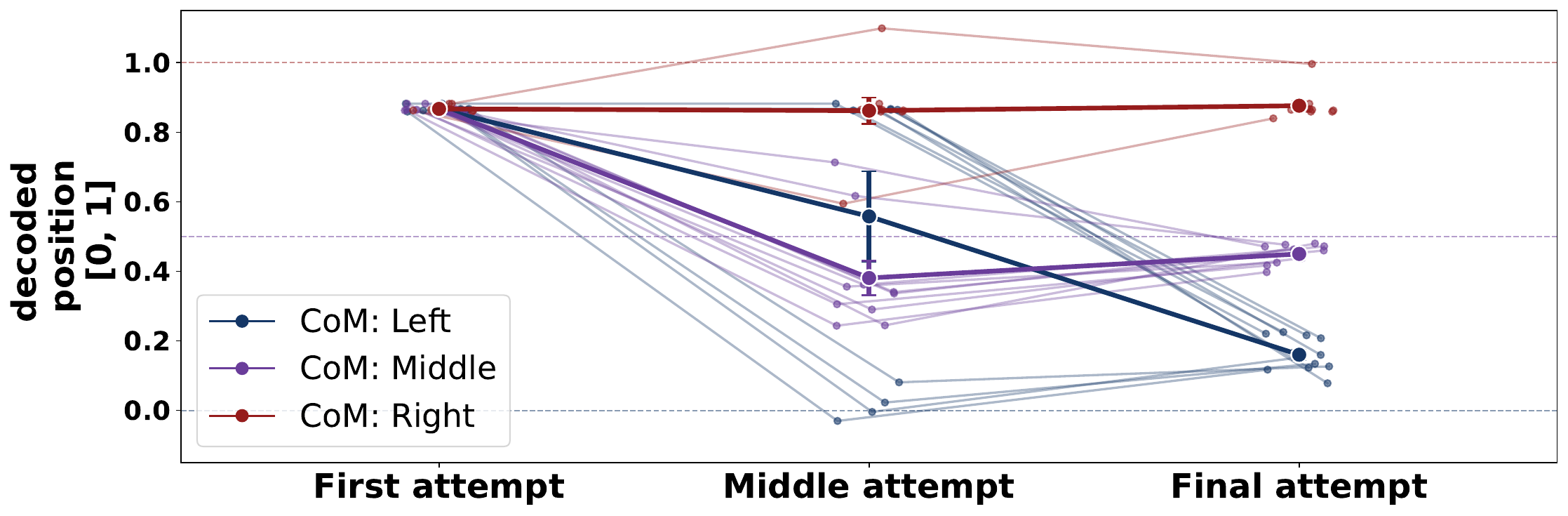}
  \caption{%
    \textbf{Decoded center of mass during real-world adaptation.} A linear probe decodes the CoM position from the belief latent at the first, middle, and final attempt of each real-world rollout; color denotes the true CoM configuration. Decoded positions start near a shared prior and separate toward their true values as adaptation proceeds.
  }
  \label{fig:real_probe}
\end{figure}

\paragraph{Belief convergence on the real robot.}
Finally, we check that the same structure holds under the sim-to-real gap. We fit a linear probe from $q$ to the normalized center-of-mass position in $[0,1]$ and apply it to the belief at each attempt of the real-world rollouts. As shown in Fig.~\ref{fig:real_probe}, the decoded center of mass starts from a common value at the first attempt, where $q = q_0$ for every rollout, and converges toward the true configuration as action outcomes accumulate. The belief update thus remains physically meaningful on real point clouds and real physics.

\newpage

\section{Real World Details}

\begin{figure}[ht]
  \centering
  \includegraphics[width=0.7\linewidth]{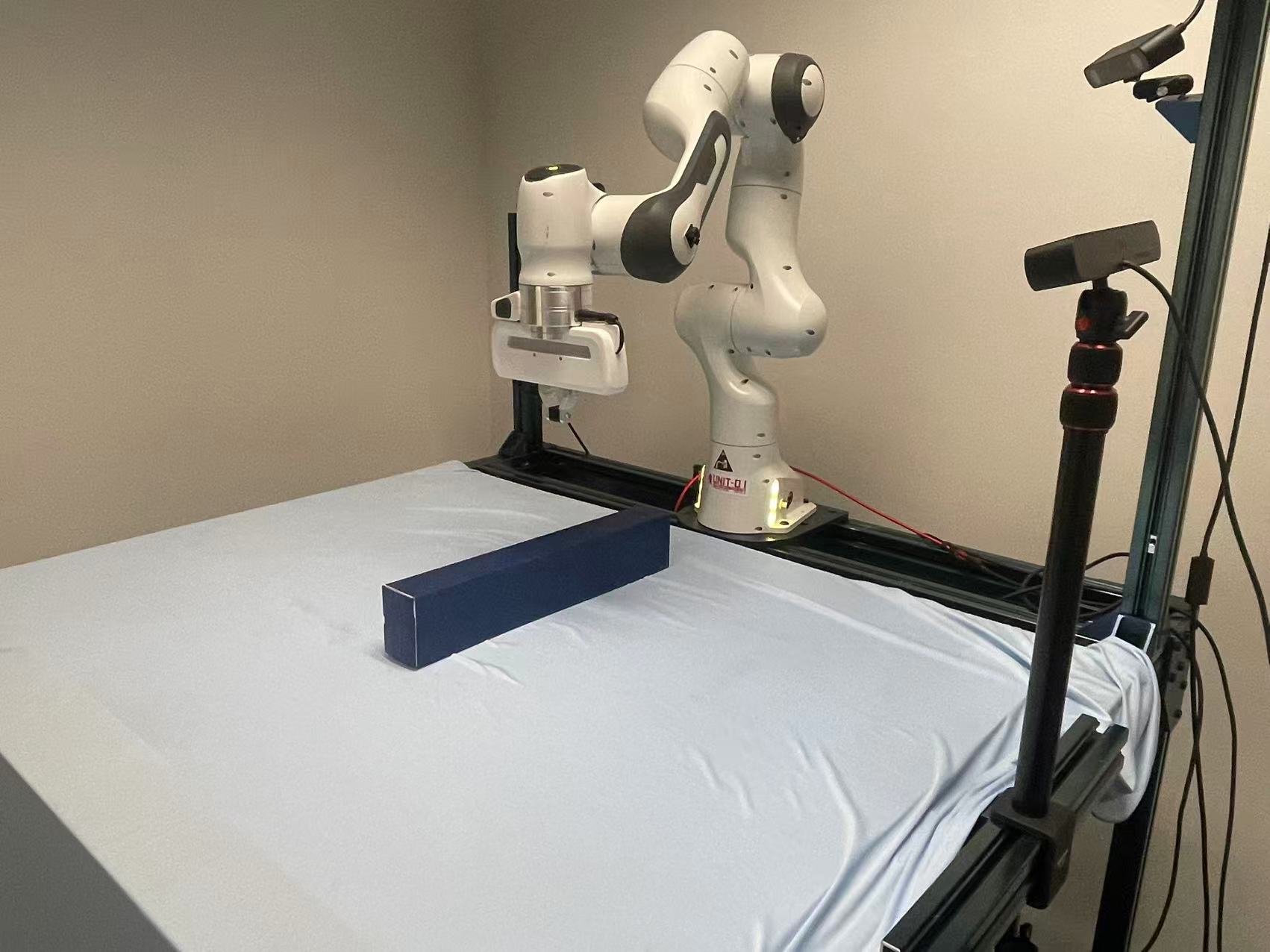}
  \caption{%
    \textbf{Real World Setup.} The Franka Emika Panda robot is equipped with a Franka Gripper and observes the workspace through a ZED camera.
  }
  \label{fig:real-world-setup}
\end{figure}

In the real-world experiments, we deploy our model, which was trained fully on simulation data, on the ISE task. The setup uses a Franka Emika Panda Robot with a Franka Gripper, and we obtain point clouds from a ZED Camera.  We use a long empty package box which we change the center of mass by putting a heavy object at different positions inside.

\end{document}